# Representing and Reasoning With Probabilistic Knowledge: A Bayesian Approach

**Marie desJardins**
SRI International
333 Ravenswood Ave.
Menlo Park CA 94025
marie@erg.sri.com

## Abstract

PAGODA (Probabilistic Autonomous GOal-Directed Agent) is a model for autonomous learning in probabilistic domains [desJardins, 1992] that incorporates innovative techniques for using the agent's existing knowledge to guide and constrain the learning process and for representing, reasoning with, and learning probabilistic knowledge. This paper describes the probabilistic representation and inference mechanism used in PAGODA.

PAGODA forms theories about the effects of its actions and the world state on the environment over time. These theories are represented as conditional probability distributions. A restriction is imposed on the structure of the theories that allows the inference mechanism to find a unique predicted distribution for any action and world state description. These restricted theories are called uniquely predictive theories. The inference mechanism, Probability Combination using Independence (PCI), uses minimal independence assumptions to combine the probabilities in a theory to make probabilistic predictions.

## 1 INTRODUCTION

PAGODA (Probabilistic Autonomous GOal-Directed Agent) is a model for autonomous learning in probabilistic domains [desJardins, 1992] that incorporates innovative techniques for using the agent's existing knowledge to guide and constrain the learning process and for representing, reasoning with, and learning probabilistic knowledge. This paper describes the probabilistic representation and inference mechanism used in PAGODA.

PAGODA's beliefs are represented as probabilistic theories about the effects of the agent's actions and the current state of the world on the environment over time. Each theory consists of a set of conditional probability distributions; each of these specifies the observed distribution of values of an output feature, given the conditioning context. Conditioning contexts consist of a perceived world and possibly an action taken by the agent. A probabilistic inference mechanism is used to make predictions about the effect of the agent's action on the theory's output feature, given the agent's perceptions (which may be the current perceived world, or a hypothetical perceived world generated by the planner). This mechanism requires determining which conditional distributions within a theory are relevant, and combining them if necessary (using minimal independence assumptions) to get a single predicted distribution.

PAGODA has been implemented in the RALPH (Rational Agent with Limited Processing Hardware) world, a simulated robot world used at UC Berkeley as a testbed for designing intelligent autonomous agents [Parr et al., 1992]. Examples from this domain are used to illustrate the work described in this paper. However, the underlying techniques are quite general and can be applied to other domains.

The rest of this paper is organized as follows: Section 2 describes PAGODA's representation for probabilistic theories. Section 3 describes the inference mechanism used to make probabilistic predictions in planning and to determine the likelihood of evidence during learning [desJardins, 1993]. Section 4 summarizes some related work on representing probabilistic beliefs, and Section 5 presents future work and conclusions.

## 2 REPRESENTING PROBABILISTIC KNOWLEDGE

PAGODA's theories are called uniquely predictive theories because a restriction is imposed on the structure of the theories that allows the inference mechanism to find a unique predicted distribution for any perceived world. The next section defines predictive theories. Uniquely predictive theories, which are a subset of predictive theories, are described in Section 2.2.

## 2.1  PREDICTIVE THEORIES

**Definition:** The **conditional probability** of $X$ given $Y$ is

$$P(X|Y) = \frac{P(X \wedge Y)}{P(Y)}$$

$X$, the **target** or **output**, and $Y$, the **conditioning context**, are first-order schemata. These schemata are required to be conjunctions of feature specifications, where each feature specification may contain internal value disjunctions, representing internal nodes in a feature value hierarchy (for example, $\texttt{vision}(t, \texttt{wall} \vee \texttt{food}, [1, 3])$ means that at time $t$, the agent sees a wall or food between 1 and 3 nodes away). In the robot domain, each schema corresponds to a set of perceived worlds. For example, the following is a valid schema in the RALPH world:

$$\texttt{vision}(t, \texttt{any-object}, 1) \wedge$$
$$\texttt{nasty-smell}(t, [10, \infty]) \wedge \Delta\texttt{u}(t, -100)$$

where $\texttt{any-object}$ represents the disjunction of all known objects. $\Delta\texttt{u}$ represents the change in utility during a time step, and is the feature that PAGODA initially forms theories to predict. Cross-feature disjunctions, such as $\texttt{vision}(t, \texttt{food}, 2) \vee \texttt{food-smell}(t, 20)$, are not allowed. Negations of features are allowed, since they may be rewritten as disjunctions.

An example of a conditional probability in the RALPH domain is

$$P(\Delta\texttt{u}(t + 1, -10)|\texttt{action}(t, \texttt{move-forward}))$$
$$= .75 \qquad (1)$$

The variable $t$ stands for any time at which this conditional probability is the most specific in the theory; that is, the knowledge we have about the situation at time $t$ implies this conditioning context and does not imply any other more specific conditioning context. Variables are *not* universally quantified, since they cannot be instantiated without examining the rest of the theory. The semantics of any individual probability within a theory will therefore depend on the content of the rest of the theory, as well as on the inference mechanism used to instantiate the variables and make predictions.

Intuitively, the meaning of Equation 1 is: given that an agent executes the action $\texttt{move-forward}$ at time $t$—and that is all the relevant information the agent has—the probability that the agent's change in utility at time $t + 1$ will be -10 is .75. The information in the conditioning context is assumed to be the only relevant information if the agent has no other probability with a more specific conditioning context. For example, if the agent also knows that $\texttt{vision}(t, \texttt{wall}, 1)$ holds, and the theory contains the conditional probability

$$P(\Delta\texttt{u}(t + 1, -10)|\texttt{action}(t, \texttt{move-forward})$$
$$\wedge \texttt{vision}(t, \texttt{wall}, 1)) = 0$$

this more relevant conditional probability will be used (and Equation 1 has no bearing on the prediction the agent makes). On the other hand, the agent may have more knowledge about $t$—such as the fact that $\texttt{nasty-smell}$ was 0—that is not mentioned in any probability in the theory; this information is considered to be irrelevant in the context of the current theory.

Using conditioning, relevance, and specificity in this way yields a quasi-non-monotonic representation: adding new knowledge (i.e., new conditional probabilities) to a theory doesn't change the *truth* of the rest of the probabilities in the theory, but it may change their range of applicability, and therefore change their semantics.

**Definition:** A **conditional distribution**, which we will usually refer to as a **rule**, is a set of $n$ conditional probabilities on a target schema $G$ (the output feature), with mutually exclusive partial variable substitutions $\theta_1 \ldots \theta_n$ and common conditioning context $C$, such that

$$\sum_{i=1}^{n} P(G\theta_i|C) = 1$$

If $C$ contains all of the relevant information for the situation about which predictions are being made, this distribution is used to predict the probability of each value of $G$.

For example, RALPH's utility goes up when it does a $\texttt{munch}$ action, but only if there is food in the same node. If this has been true on half of the occasions it's tried a $\texttt{munch}$ action, it may have a rule predicting $\Delta\texttt{u}(t1, du)$ that contains the probabilities

$$P(\Delta\texttt{u}(t + 1, 90)|\texttt{action}(t, \texttt{munch})) = .5$$
$$P(\Delta\texttt{u}(t + 1, -10)|\texttt{action}(t, \texttt{munch})) = .5 \quad (2)$$

A rule with an empty conditioning context is referred to as a **prior distribution** on $G$, or a **default rule** for $G$. A **prediction** on $G$ is the set of probabilistic outcomes specified by a conditional distribution. The rule in Equation 2 makes the prediction

$$\{(\Delta\texttt{u}(t + 1, 90), .5), (\Delta\texttt{u}(t + 1, -10), .5)\}$$

**Definition:** A **predictive theory** on output feature $G$ is a set of $m$ conditional distributions, or rules, on $G$, with conditioning contexts $C_1 \ldots C_m$ (which must be distinct but not necessarily disjoint), such that any *situation* (consisting of a perceived world and possibly an action) implies at least one of the conditioning contexts.

As long as a set of distinct rules on a output feature includes a default rule it is guaranteed to be a predictive theory.

A predictive theory stores all of the beliefs the agent has about the output feature $G$. The rules in a theory are indexed by their conditioning contexts (i.e., the situations in which they apply). Using a specificity relation between conditioning contexts, the rules can be organized into a DAG in which a child is always more specific than its parents. A rule in the theory may have multiple parents, but no rule may be an ancestor of itself.

Figure 1 shows an example of a predictive theory on a output feature $G$, drawn as a DAG. Only the conditioning contexts are shown, indicating the structure of the theory. A conditional distribution is actually stored at each node. For example, the bottom node represents a rule containing conditional probabilities of the form $P(G\theta_i|A(x) \wedge B(x)) = p_i$.

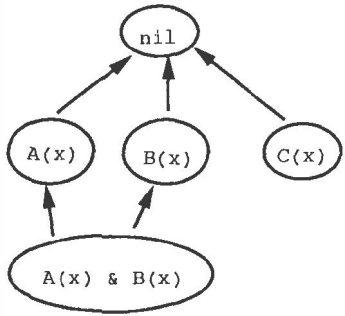

Figure 1: Example of a predictive theory

## 2.2   UNIQUELY PREDICTIVE THEORIES

A predictive theory may correspond to many different complete probability distributions. In principle, probabilities that are not specified by the theory may take on any value that is consistent with the probabilities in the theory. For example, given only the probabilities in Figure 1, $P(G|D(x))$ may take on any value. In order to make predictions about situations that are not explicitly mentioned as conditioning contexts in a rule in the theory, a single distribution must be found that specifies the remaining probabilities.

The Maximum Entropy (ME) principle [Levine and Tribus, 1979] provides one method for finding a "best" distribution, using the rules in a theory as constraints on the distribution. The distribution chosen using this method will add the least information possible to the existing theory. However, in the general case (i.e., for arbitrary constraints), ME is intractable.

A less expensive approach is to identify reasonable independence assumptions and use them to find the joint distribution. We restrict the set of allowed theories so that a unique distribution can be found using only simple independence assumptions that are consistent with the theory.

If the induction mechanism finds a theory that contains all dependencies that actually exist and no oth-ers, it is safe in the limit to assume that any dependence not represented in the agent's theory does not exist. PAGODA's Bayesian learning method will discard any theory that contains additional dependencies (irrelevant rules) in favor of a simpler theory without the irrelevant rules; similarly, any theory that is missing dependencies that actually exist (i.e., statistically significant correlations in the data) will be discarded for one that contains the dependencies.

PAGODA's inference mechanism is based on this independence-assumption approach. The technique involves finding a set of most specific rules that apply to the situation of interest. Shared features in the conditioning contexts of these rules are identified; it is assumed that the remaining features are independent, given the shared features.

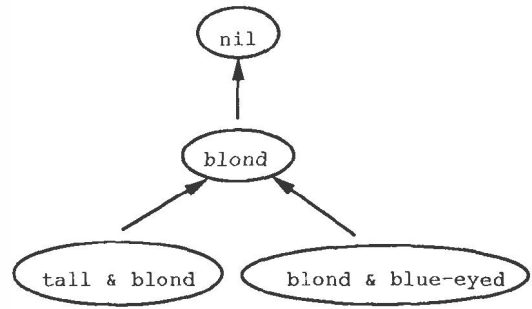

Figure 2: Sample uniquely predictive theory

The theory shown in Figure 2 specifies the conditional distributions corresponding to

$$P(\mathtt{Swedish}(x))$$
$$P(\mathtt{Swedish}(x)|\mathtt{blond}(x))$$
$$P(\mathtt{Swedish}(x)|\mathtt{tall}(x) \wedge \mathtt{blond}(x))$$
$$P(\mathtt{Swedish}(x)|\mathtt{blond}(x) \wedge \mathtt{blue\text{-}eyed}(x))$$

Recall that the output feature, $\mathtt{Swedish}(x)$, does not appear in the diagram, but each rule represented will make some prediction about the probability of $\mathtt{Swedish}(x)$, given that the conditioning context holds.

Now we observe someone who is tall, blond, and blue-eyed. Given our theory, we wish to find the probability that they are Swedish, i.e.,

$$P(\mathtt{Swedish}(x)|\mathtt{tall}(x) \wedge \mathtt{blond}(x) \wedge \mathtt{blue\text{-}eyed}(x))$$

If we assume that $\mathtt{blue\text{-}eyed}$ and $\mathtt{tall}$ are independent, and that they are conditionally independent given $\mathtt{blond}$ or $\mathtt{blond} \wedge \mathtt{Swedish}$, this can be rewritten as

$$P(\mathtt{Swedish}(x)|\mathtt{tall}(x) \wedge \mathtt{blond}(x)) \, *$$
$$P(\mathtt{Swedish}(x)|\mathtt{blond}(x) \wedge \mathtt{blue\text{-}eyed}(x)) \, /$$
$$P(\mathtt{Swedish}(x)|\mathtt{blond}(x))$$

The most specific rules for this prediction are those with conditioning contexts $\mathtt{tall} \wedge \mathtt{blond}$ and $\mathtt{blond} \wedge$

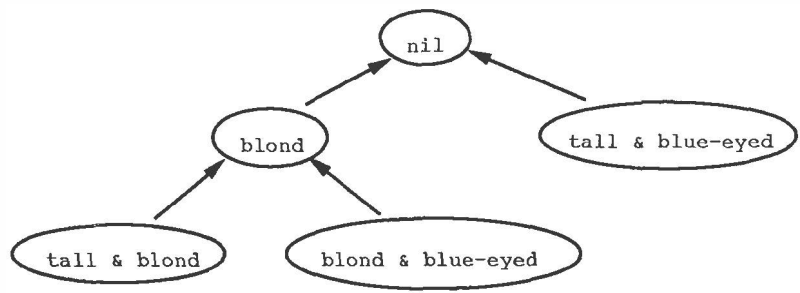

Figure 3: Unacceptable uniquely predictive theory

**blue-eyed. blond** is the shared feature of the conditioning contexts of these rules, and is used to separate their effects. The denominator represents the combined effects of the two rules; the numerator represents the overlap (essentially the shared part of the world state that was included twice). If $P(\texttt{Swedish}(x)|\texttt{blond}(x))$ were removed from the theory, we would assume that it was equal to the prior $P(\texttt{Swedish}(x))$ (which is the most specific rule for **blond**$(x)$). The complete derivation for the general case is given in Section 3.

However, if we add the distribution specifying

$$P(\texttt{Swedish}(x)|\texttt{tall}(x) \land \texttt{blue-eyed}(x))$$

yielding the theory shown in Figure 3, we would need to assume that **blond**, **blue-eyed**, and **tall** were all independent. But if they were, this wouldn't be the simplest theory: a perfect induction mechanism would have preferred the theory shown in Figure 4.

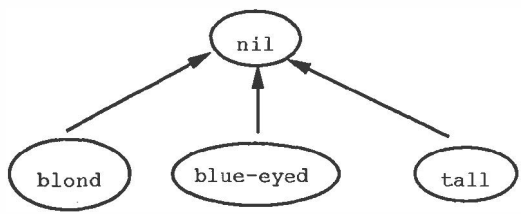

Figure 4: Preferred uniquely predictive theory

The inference mechanism does not work on theories such as the one in Figure 3, which have interlinked dependencies such that independent features cannot be pulled out individually. In an inductive learning environment, this is acceptable in principle, since such theories will not be generated by an ideal learning mechanism. However, in practice, such theories are sometimes generated. Therefore, although the formal theory does not cover this case, PCI includes heuristics for handling it; these are described in Section 3. We now formalize the restriction that excludes theories such as the one shown in Figure 3.

> **Definition:** The **most specific rules** (MSRs) for a situation $S$ are the rules in the theory whose conditioning contexts are more general than $S$, such that no more specific rule's conditioning context is also more general than $S$.

A rule is an MSR for a situation $S$ if its conditioning context is more general than $S$ and it has no children whose conditioning contexts are also more general than $S$. The MSRs for $S$ are the conditional distributions that will be used to make predictions about the outcome of the specified action in the world state. Uniquely predictive theories, described in the next section, are a restricted form of predictive theories that allow MSRs to be combined using simple independence assumptions; the inference mechanism which does this is described in Section 3.

> **Definition:** A set of rules in a theory is a **valid set of MSRs** if it corresponds to some situation; i.e., there must be a situation (perceived world plus an action) that would have the set of rules as its MSRs.

In Figure 1, invalid sets of MSRs include $\{\texttt{nil}, \texttt{C}(x)\}$ ($\texttt{C}(x)$ should be the only MSR) and $\{\texttt{A}(x), \texttt{B}(x)\}$ (since $\texttt{A}(x) \land \texttt{B}(x)$ would be a valid MSR for the situation).

> **Definition:** The **shared features** of a set of rules are the features that appear in all of the conditioning contexts and have some value in common.

"Shared features" may also refer to this shared set of values for the features, in which case they may be thought of as the minimum specializations of the common features. For example, the shared feature of $\texttt{blond}(x) \land \texttt{blue-eyed}(x)$ and $\texttt{tall}(x) \land \texttt{blue-eyed}(x)$ is $\texttt{blue-eyed}(x)$. The shared features of

$$\texttt{vision}(10, \texttt{food}, 2) \land \texttt{food-smell}(10, 5)$$
$$\land\, \texttt{nasty-smell}(10, 10) \land \Delta u(10, -10)$$

and

$$\texttt{vision}(t, \texttt{food}, [1, 3]) \land \texttt{action}(t, \texttt{move-forward})$$
$$\land\, \texttt{food-smell}(t, 20)$$

are

$$\texttt{vision}(t, \texttt{food}, 2) \land \texttt{action}(t, \texttt{move-forward})$$

The action appears in the shared features because the first situation implicitly allows any value for the action, so the shared value is `move-forward`.

> **Definition:** A set of rules is **separable** if there is some rule in the set (which is also referred to as separable with respect to the rest of the set) whose conditioning context can be split into two parts: one group of features that is shared with a single other rule in the set, and one group of features that is shared with no other rule in the set. Either group of features may be empty.

The restriction on uniquely predictive theories is simply that every valid set of MSRs must be separable. Figure 3 violates this restriction because the valid set of MSRs

$$\{\texttt{tall}(x) \wedge \texttt{blond}(x), \texttt{blond}(x) \wedge \texttt{blue-eyed}(x),$$
$$\texttt{blue-eyed}(x) \wedge \texttt{tall}(x)\}$$

is not separable: all of the rules in the set share features with both of the other rules.

# 3    PROBABILISTIC INFERENCE

This section describes Probability Combination using Independence (PCI), the inference method that is applied to a theory $T$ to compute the distribution of $T$'s output feature $G$, given a situation $S$. Given a set of MSRs for $S$, PCI iteratively finds a separable rule in the set, computes its contribution to the overall probability using independence assumptions, and recurses using the remaining rules as the new set of MSRs to explain the remaining features. The algorithm operates as follows:

1. Let $R$ be the set of the $n$ most specific rules (MSRs) in $T$ that apply to $S$. This set consists of all rules, $r_i$, whose conditioning context $C_i$ is satisfied by the situation, where no strictly more specific rule also satisfies the situation:

$$R = \{r_i : [S \to C_i] \wedge$$
$$\neg \exists r_k, k \neq i : [(S \to C_k) \wedge (C_k \to C_i)]\}$$

2. The rules are ordered so that each rule $r_i$ is separable given the set of rules $r_{i+1}, \ldots, r_n$. Recall that $r_i$ is separable with respect to a set of rules if its conditioning context can be split into two parts: $f_i^s$, a group of features (possibly empty) that is shared with some rule in the set, and $f_i^u$, the remaining features, which are shared with no other rule in the set (i.e., are unique to $r$ in this set of rules). This is guaranteed to be possible if $T$ is a uniquely predictive theory, since each set of rules $r_i \ldots r_n$ is a valid set of MSRs.

3. The probability of $G\theta$ is computed, for each $\theta$ common to all rules in the set of MSRs (i.e., for values of $G$ that are assigned non-zero probability by every rule in the set). If we assume that $f_i^u$ is independent of the features only found in the rest of the rules (i.e., of $\bigcup_{k>i} C_k - f_i^s$), and also conditionally independent of those features given $G$ and $f_i^s$ (yielding a total of $2(n-1)$ independence assumptions, all consistent with the dependencies explicitly expressed in the theory), this probability is equal to

$$P(G\theta|S) = \frac{\prod_{i=1}^{n} P(G\theta|C_i)}{\prod_{j=1}^{n-1} P(G\theta|f_j^s)} \qquad (3)$$

(The derivation of this equation is given below.) If $n$ is 1, the product in the denominator is defined to be 1, and the predicted distribution on an output feature when only one rule applies is simply the distribution given by that rule.

4. The probabilities in the denominator of Equation 3 are computed by applying PCI recursively.

The resulting probabilities are derived probabilities, which may be used to make further inferences in the planning process, but otherwise are not reused. Specifically, they are not stored in the theory. This keeps the empirical probabilities represented in the theory distinct from the inferred, subjective probabilities (they are subjective because the independence assumptions have not been directly validated against the data).

The formula given in Equation 3 is derived as follows. Consider the effects of pulling out the first MSR, $r_1$, and assuming that its unique features $f_1^u$ are independent of the remaining features ($\bigcup_{j>1} C_j - f_1^s$), and independent of these features given $G$ and $f_1^s$. In order to simplify the derivation somewhat, we assume that $r_2$ is the rule that shares the feature $f_1^s$. This is not necessarily the case: in fact, $r_2$ is simply the next separable rule. However, making this assumption does not affect the validity of the derivation. We will refer to the features in $r_2$ that are not shared with $r_1$ as $f_2^u$. Then using Bayes' rule[1] gives us the derivation in Figure 5. Iterating on the last term in the numerator yields Equation 3.

If the inductive learning algorithm is "perfect"—i.e., it identifies all dependencies that exist—this procedure will be guaranteed to work, because the independence assumptions will be correct. However, in practice, theories are often not perfect, due to limited data or an inadequate search heuristic. The result is that the procedure may not yield a valid distribution on $G$: the computed probabilities may sum to less than or more than one. In this case, we normalize the probabilities to sum to 1 and proceed as usual. In the extreme case, the sum of the probabilities will be zero if every

---

[1]This is a slightly non-standard version of Bayes' rule. The general form of the rule we use here is:

$$P(X|Y \wedge K) = \frac{P(X|K)P(Y|X \wedge K)}{P(Y|K)}$$

$$P(G|S) = P(G|f_1^u \wedge f_1^s \wedge f_2^x \wedge C_3 \ldots C_n)$$

$$= \frac{P(G|f_1^s)\, P(f_1^u \wedge f_2^x \wedge C_3 \ldots C_n | G \wedge f_1^s)}{P(f_1^u \wedge f_2^x \wedge C_3 \ldots C_n | f_1^s)}$$

$$= \frac{P(G|f_1^s)\, P(f_1^u | G \wedge f_1^s)\, P(f_2^x \wedge C_3 \ldots C_n | G \wedge f_1^s)}{P(f_1^u | f_1^s)\, P(f_2^x \wedge C_3 \ldots C_n | f_1^s)}$$

$$= \frac{P(G|f_1^s)\frac{P(f_1^u | f_1^s)\, P(G|f_1^u \wedge f_1^s)}{P(G|f_1^s)}\frac{P(f_2^x \wedge C_3 \ldots C_n | f_1^s)\, P(G|f_2^x \wedge C_3 \ldots C_n \wedge f_1^s)}{P(G|f_1^s)}}{P(f_1^u | f_1^s)\, P(f2 \wedge C_3 \ldots C_n | f_1^s)}$$

$$= \frac{P(G|f_1^u \wedge f_1^s)\, P(G|f_1^s \wedge f_2^x \wedge C_3 \ldots C_n)}{P(G|f_1^s)}$$

$$= \frac{P(G|C_1)\, P(G|C_2 \ldots C_n)}{P(G|f_1^s)}$$

Figure 5: Derivation of Combined Probability

value of the output feature is assigned zero probability by some MSR. In this case, PCI assumes that not enough data has been collected to cover the current case adequately, and uses the less specific probability $P(G|f_*^s)$, where $f_*^s$ is the set of features that are shared by all MSRs (possibly empty, in which case the prior probability $P(G)$ is used).

### 3.1  AN EXAMPLE OF PCI

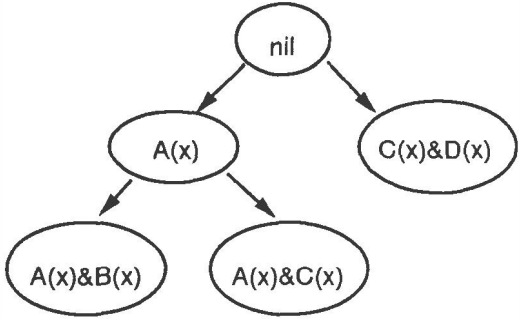

Figure 6: Theory to be used for making predictions

Taking the theory represented in Figure 6 as a predictive theory on a Boolean feature $G$, and leaving out the argument $x$, the theory can be rewritten as a set of conditional probabilities: The theory in Figure 6 represents the probabilities $p_g$, $p_a$, $p_{ab}$, $p_{ac}$, and $p_{cd}$.

$$
\begin{aligned}
R_g: \quad & p_g = P(G) \\
R_a: \quad & p_a = P(G|A) \\
R_{ab}: \quad & p_{ab} = P(G|A \wedge B) \\
R_{ac}: \quad & p_{ac} = P(G|A \wedge C) \\
R_{cd}: \quad & p_{cd} = P(G|C \wedge D)
\end{aligned}
$$

In order to find any probability which is not explicitly represented in the theory, PCI must be applied. The simplest case is when only one rule applies to the new probability. For example, for the situation $A \wedge D$, $R$

(the set of most specific rules) is just $\{R_a\}$, so

$$P(G|A \wedge D) = P(G|A) = p_g$$

For the situation $A \wedge B \wedge C \wedge D$, $R$ is less $\{R_{ab}, R_{ac}, R_{cd}\}$. $R_{ac}$ is not separable given $R_{ab}$ and $R_{cd}$, since it shares the feature $A$ with $R_{ab}$ and $C$ with $R_{cd}$. $R_{ab}$ is separable given $R_{ac}$ and $R_{cd}$, since it only shares features with $R_{ac}$, and $R_{ac}$ is separable given $R_{cd}$, so a valid ordering is $R = (R_{ab}, R_{ac}, R_{cd})$. Applying Equation 3 gives

$$P(G|A \wedge B \wedge C \wedge D) =$$
$$\frac{P(G|A \wedge B)\, P(G|A \wedge C)\, P(G|C \wedge D)}{P(G|A)\, P(G|C)}$$

$P(G|C)$ must be computed recursively: in this case, $R$ is $\{R_g\}$, so $P(G|C) = P(G)$ and

$$P(G|A \wedge B \wedge C \wedge D) =$$
$$\frac{P(G|A \wedge B)\, P(G|A \wedge C)\, P(G|C \wedge D)}{P(G|A)\, P(G)}$$

## 4  RELATED WORK

In order to use probabilistic knowledge in an automated learning system, a formal system for representing and reasoning with probabilities is required. In particular, given a set of generalized conditional probabilities (i.e., a probabilistic theory) and some (possibly probabilistic) knowledge about a particular object, the system must be able to make probabilistic predictions about unobserved properties of the object.

Kyburg [1974] defined the reference class for a proposition as the features that are relevant for making probabilistic predictions about the proposition. For example, the appropriate reference class for determining whether or not Chilly Willy can fly in the previous example is the class of penguins. The reference class for a proposition will depend on what is being predicted and on what probabilities are represented in the theory or set of beliefs. Once the reference class is found, determining the probability of the proposition may require probabilistic inference from the beliefs in the theory.

Bacchus's [1990] probabilistic logic and Pearl's [1988b] belief nets provide formalisms for representing probabilistic knowledge. We discuss these two approaches in the following sections.

## 4.1  LOGIC AND PROBABILITY

Bacchus's [1990] probabilistic logic is a formal language for representing probabilistic knowledge using first-order logic. The language provides a representation for both statistical probabilities (defined in terms of observed frequencies of events) and subjective probabilities (degrees of belief derived from the statistical probabilities). The inference mechanism provides for some manipulation of the statistical probabilities using standard axioms of probability, and for direct inference from statistical to subjective probabilities using the narrowest reference class.

The subjective probability of a proposition is given a formal interpretation as the total probability mass of all possible worlds in which the proposition is true. An example (given by Bacchus) of a subjective probability in the language is "birds fly with probability at least 0.75," written as

$$\forall x.\text{prob}(\text{bird}(x)) > 0 \rightarrow \text{prob}(\text{fly}(x)|\text{bird}(x)) > 0.75$$

The antecedent is necessary because Bacchus does not permit conditioning on a statement which is known to be false. Qualitative relationships between probabilities can also be expressed; for example, conditional independence can be explicitly written as

$$\text{prob}(A \wedge B|C) = \text{prob}(A|C)\,\text{prob}(B|C)$$

Statistical probabilities, representing frequencies of events in actual trials, have a different syntax, and require "placeholder variables" to indicate which variables are intended to vary randomly. For example, the statement "ten tosses of a coin will land heads 5 times with greater than 95% probability" is written as

$$[\text{freq-heads}(x) = .5|10\text{-tosses}(x)]_x > 0.95 \quad (4)$$

Direct inference from statistical to subjective probabilities is based on finding a statistical probability with the same reference class as the desired subjective probability. If no such probability is available, a simple type of independence is assumed nonmonotonically, and the "next narrowest" reference class for which a probability is available is used. For example, if one wishes to find the probability that a particular sequence of 10 tosses of a quarter will yield five heads, and the only statistical probability available is Equation 4, the direct inference mechanism non-monotonically assumes independence of freq-heads and quarter, given 10-tosses, yielding

$$\text{prob}(\text{freq-heads}(T)|10\text{-tosses}(T) \wedge \text{quarter}(T))$$
$$= [\text{freq-heads}(x) = .5|10\text{-tosses}(x) \wedge \text{quarter}(x)]_x$$
$$= [\text{freq-heads}(x) = .5|10\text{-tosses}(x)]_x$$
$$> 0.95$$

While Bacchus's language provides a useful formalism for representing many aspects of probabilistic reasoning, including certain forms of default reasoning, it does not provide a representation for beliefs about relevance, nor does it allow default assumptions such as independence or maximum entropy to be used in the inference process.

## 4.2  BELIEF NETWORKS

A belief network is a compact representation of a complete joint probability distribution on a set of propositions. Each proposition is represented as a node, and conditional probabilities (dependencies) are represented as links between nodes. Any nodes that are not directly connected are assumed to be conditionally independent, given the intervening nodes.

A probability matrix is stored at each node in the network, representing the conditional probability distribution for that node given its set of parent nodes. The joint probability distribution $P(x_1, \ldots, x_n)$ for the $n$ nodes in a belief network is the product of the conditional probabilities of all nodes given their parents.

One problem with belief nets as presented above is that they require a probability matrix of size $k_i \prod_{j \in \text{parents}} k_j$ at every node $i$ (where $k_i$ is the number of values that the random variable at node $i$ takes). Pearl [1988a] gives several models for computing this matrix from a subset of the probabilities; he refers to these models as Canonical Models of Multicausal Interaction (CMMIs). The noisy-OR model of disjunctive interaction models a set of independent causes (parents) of an event (node). Each cause has an associated "exception"— a random variable which, if true, will inhibit the effect of the cause on the event. For example, Pearl gives a situation where the event in question is a burglar alarm going off; the two causes are a burglar and an earthquake; and the two inhibitors are that the burglar is highly competent and that the earthquake has low vertical acceleration. Given an event $E$ with Boolean causes $C_i$ and associated exceptions with probabilities $q_i$, the overall probability of the event is given as:

$$P(E|c_1, \ldots, c_n) = \prod_{i:c_i isTRUE} q_i$$

This model allows the probability matrix to be computed from only $n$ probabilities, instead of the $2^n$ that would be required to enumerate all of the conditional probabilities in the complete matrix.

PAGODA's uniquely predictive theories are a hybrid of rule-based approaches and the belief-net method of representing dependencies. They consist of rules, which are easy to manipulate, perform inference with, and learn using familiar and intuitive inference rules and inductive operators. However, the rules are not modular: the semantics does not allow the inference rules to be applied to a theory without knowing what other rules exist in the system.

PCI provides the equivalent of a sophisticated CMMI for a node in a belief network. The probabilities stored in PAGODA's theories are used to compute the entries that would appear in the probability matrix dynamically, assuming independence where necessary. Therefore, PCI could be used within a belief net framework to reduce the number of probabilities that must be precomputed and stored at each node.

## 5    CONCLUSIONS

We have described uniquely predictive theories and PCI, a representation and inference mechanism for predictive probabilistic theories. PAGODA, a model for autonomous learning [desJardins, 1992], uses PCI for representing learned theories, for evaluating potential theories, and for making probabilistic predictions for planning. The implementation of PAGODA in the RALPH world, a simulated robot domain, has shown uniquely predictive theories and PCI to be a useful and powerful mechanism for representing probabilistic predictive theories.

The constraints on theories allow certain kinds of independence to be captured automatically, but it may be desirable to allow more complex interactions. One way to do this would be to identify more common types of interactions and provide general solutions for computing the effects of those interactions (as we have already done for independence).

Another open area for research is using the results of the inference process to guide future learning by identifying weaknesses with the existing theory. For example, the cases described in Section 3 (when the distribution yielded by PCI is invalid, requiring normalization or the use of a less specific probability) indicate that something is wrong with the learned theories: either an important dependence is not captured, or the probabilities are wrong. Representing the confidence PAGODA has in its theories (e.g., by maintaining second-order probabilities on the rules) would provide useful information for the system to determine whether a problem actually exists and, if so, where it lies.

Uniquely predictive theories and PCI provide a powerful new mechanism for representing and reasoning with probabilistic information, complementing previous work in the areas of probabilistic logics and belief networks.

### References

[Bacchus, 1990] Fahiem Bacchus. *Representing and Reasoning with Probabilistic Knowledge: A Logical Approach to Probabilities.* MIT Press, 1990.

[desJardins, 1992] Marie desJardins. *PAGODA: A Model for Autonomous Learning in Probabilistic Domains.* PhD thesis, UC Berkeley, 1992.

[desJardins, 1993] Marie desJardins. Bayesian theory evaluation: A probabilistic approach to concept learning, 1993. Submitted to MLC-93.

[Kyburg, 1974] Henry E. Kyburg. *The Logical Foundations of Statistical Inference.* Reidel, 1974.

[Levine and Tribus, 1979] Raphael D. Levine and Myron Tribus, editors. *The Maximum Entropy Formalism.* MIT Press, 1979.

[Parr *et al.*, 1992] Ronald Parr, Stuart Russell, and Mike Malone. The RALPH system. Technical report, UC Berkeley, 1992. (Forthcoming).

[Pearl, 1988a] Judea Pearl. On logic and probability. *Computational Intelligence*, 4(1):99–103, 1988.

[Pearl, 1988b] Judea Pearl. *Probabilistic Reasoning in Intelligent Systems: Networks of Plausible Inference.* Morgan Kaufmann, 1988.